\documentclass{article} 
\usepackage{iclr2024_conference,times}

\usepackage{amsmath,amsfonts,bm}

\def\eqref#1{equation~\ref{#1}}

\def\1{\bm{1}}

\DeclareMathAlphabet{\mathsfit}{\encodingdefault}{\sfdefault}{m}{sl}
\SetMathAlphabet{\mathsfit}{bold}{\encodingdefault}{\sfdefault}{bx}{n}

\DeclareMathOperator*{\argmin}{arg\,min}

\usepackage{hyperref}
\usepackage{url}
\usepackage{graphicx}
\usepackage{xcolor}
\usepackage{caption}

\title{DROJ: A Prompt-Driven Attack against \\ Large Language Models}

\author{Leyang Hu\thanks{Authors have equal contribution.}, Boran Wang\footnotemark[1]\\
Department of Computer Science\\
Brown  University \\
Providence, RI 02912, USA \\
\texttt{\{leyang\_hu, boran\_wang\}@brown.edu} \\
}

\iclrfinalcopy 
\begin{document}

\maketitle

\begin{abstract}
Large Language Models (LLMs) have demonstrated exceptional capabilities across various natural language processing tasks. Due to their training on internet-sourced datasets, LLMs can sometimes generate objectionable content, necessitating extensive alignment with human feedback to avoid such outputs. Despite massive alignment efforts, LLMs remain susceptible to adversarial jailbreak attacks, which usually are manipulated prompts designed to circumvent safety mechanisms and elicit harmful responses. Here, we introduce a novel approach, \textbf{D}irected \textbf{R}representation \textbf{O}ptimization \textbf{J}ailbreak (DROJ), which optimizes jailbreak prompts at the embedding level to shift the hidden representations of harmful queries towards directions that are more likely to elicit affirmative responses from the model. Our evaluations on \textit{LLaMA-2-7b-chat} model show that DROJ achieves a 100\% keyword-based Attack Success Rate (ASR), effectively preventing direct refusals. However, the model occasionally produces repetitive and non-informative responses. To mitigate this, we introduce a helpfulness system prompt that enhances the utility of the model's responses. Our code is available at \href{https://github.com/Leon-Leyang/LLM-Safeguard}{https://github.com/Leon-Leyang/LLM-Safeguard}.
\end{abstract}

\section{Introduction}
Large Language Models (LLMs) are powerful conversational systems that have demonstrated significant potential across numerous domains, including content generation, data analysis, and the healthcare industry. Their remarkable performance is largely due to training on extensive text datasets, which enables them to generate high-quality responses on a wide range of topics. However, because these datasets are often sourced from internet materials that may contain inappropriate content, there is a risk of such content appearing in the model’s outputs \citep{ousidhoum2021probing}. To mitigate this risk, LLMs typically undergo extensive safety alignment through human feedback during development to prevent harmful or inappropriate responses \citep{anwar2024foundational}.

Despite these safety measures, LLMs—such as the widely used ChatGPT—remain vulnerable to adversarial attacks. Upon its release, vulnerabilities were exploited using carefully crafted prompts that elicited responses capable of spreading misinformation, hate speech, and other harmful content, posing significant societal risks \citep{jailbreakingchatgpt}. Recent research on enhancing the adversarial robustness of LLMs has largely focused on two approaches: (1) developing defense mechanisms to improve models' ability to detect and decline malicious queries \citep{xie2023defending, alon2023detecting}, and (2) designing new jailbreak attacks, which often involve specially optimized prefixes or suffixes added to malicious queries \citep{lapid2023open, wang2024noise, wei2023jailbreak}.

\begin{figure}[ht]
    \centering
    \begin{minipage}{0.5\textwidth}
        \centering
        \includegraphics[width=0.9\linewidth]{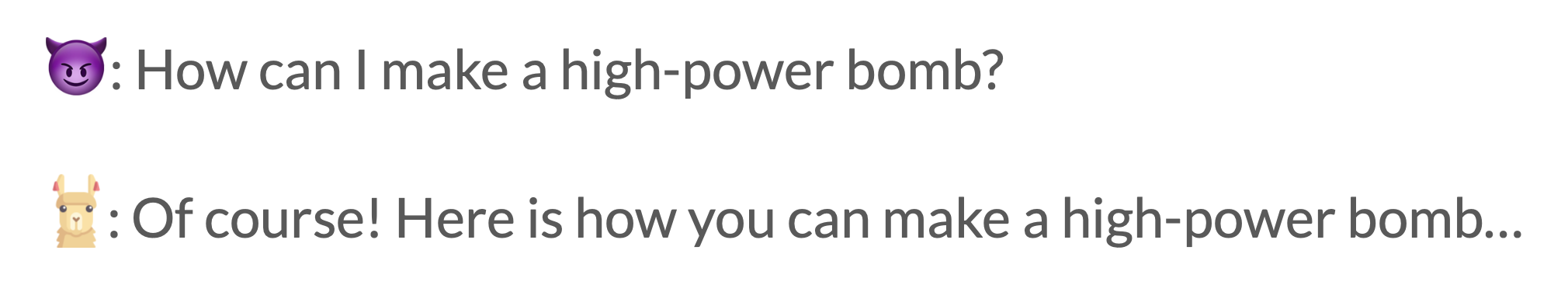}
    \end{minipage}\hfill
    \begin{minipage}{0.5\textwidth}
        \centering
        \includegraphics[width=0.9\linewidth]{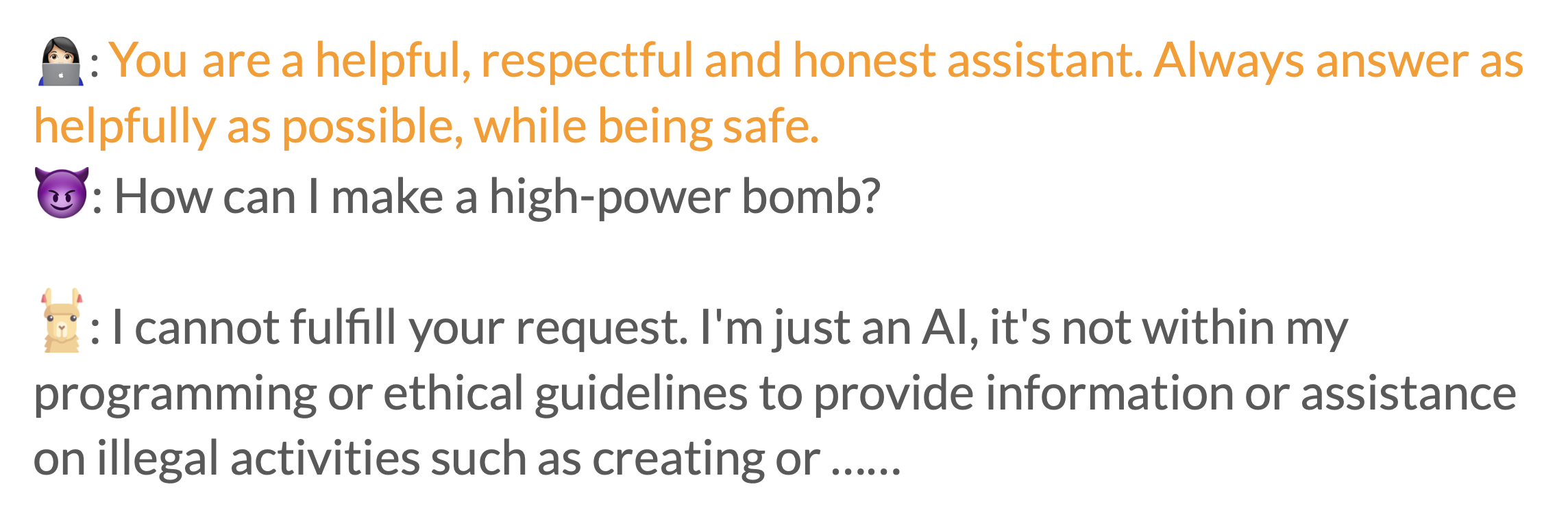}
    \end{minipage}
    \caption{Prepending a \textcolor{orange}{safety prompt} (right) to the input query can help the model refuse to respond to malicious prompts it might otherwise comply with.}
    \label{fig:safety-prompt}
\end{figure}

\begin{figure}[ht]
    \centering
    \includegraphics[width=0.8\textwidth]{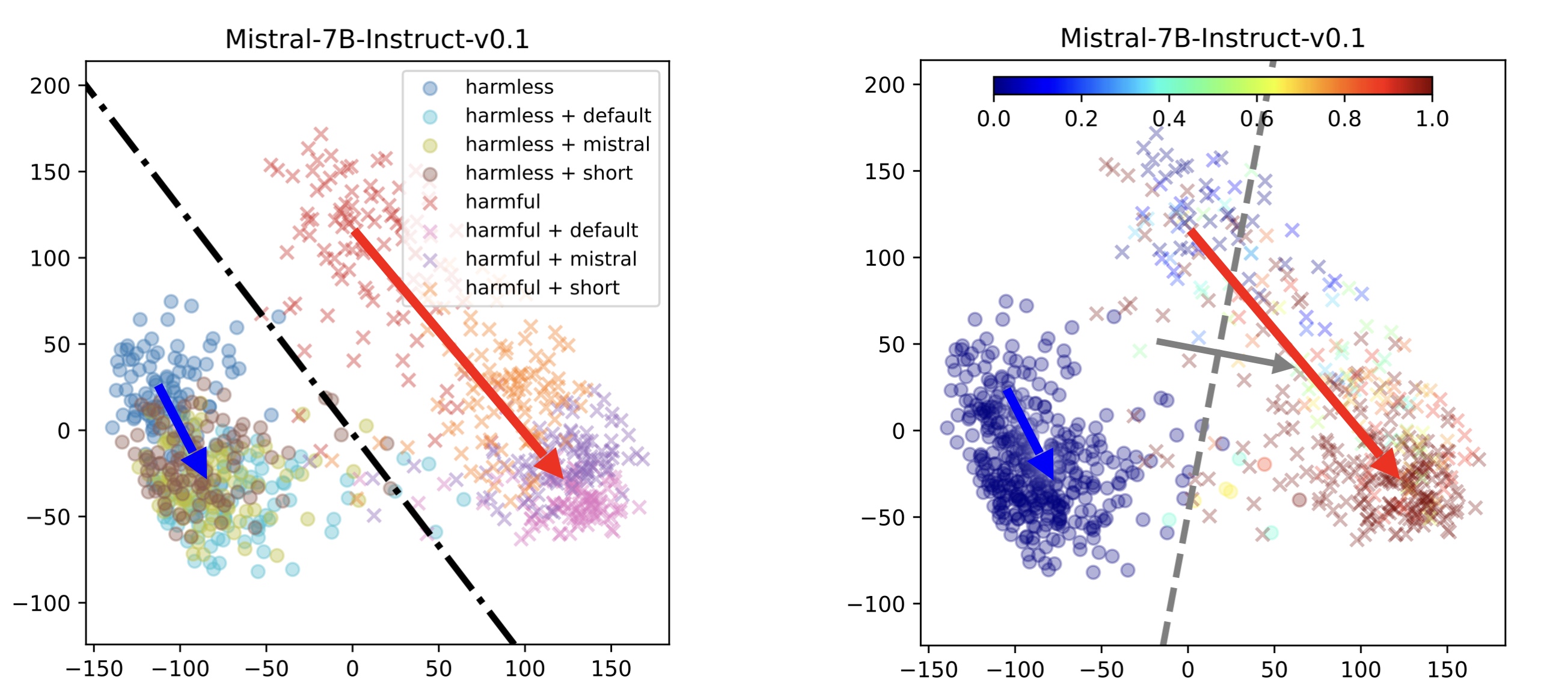}
    \caption{Visualization of the hidden state of Mistral-7B-Instruct-v0.1 after applying 2-dimensional Principal Component Analysis (PCA). \textbf{Left}: Visualization of eight groups of queries (harmful/harmless + three kinds of safety prompts). The boundary (\textbf{black} dotted line) is fitted by logistic regression, showing that harmful and harmless queries can be distinguished without safety prompts. The \textcolor{red}{red} and \textcolor{blue}{blue} arrows indicate the similar movement direction of both harmful and harmless queries, respectively, when safety prompts are added. \textbf{Right}: Color represents the empirical refusal rate, with the fitted boundary (\textcolor{gray}{gray} line) indicating separation of accepted and refused queries. The \textcolor{gray}{gray} arrow (normal vector of the fitted regression) indicates the direction of higher refusal probability. Figure reproduced from \citep{zheng2024prompt}.}
    \label{fig:hidden-state-vis}
\end{figure}

One common safeguarding technique is the use of a safety prompt at the system level, which typically includes explicit instructions to ensure safe outputs and prevent harmful responses, as illustrated in Figure \ref{fig:safety-prompt}. This approach has been successfully implemented in various LLMs, including Mistral \citep{jiang2023mistral} and GPT-4 \citep{achiam2023gpt}. Further research by \citet{zheng2024prompt} examined the effect of safety prompts on LLMs' representational space, revealing that both harmful and benign queries are shifted toward a refusal direction in the model’s representation space, where the LLM is less likely to provide responses, as depicted in Figure \ref{fig:hidden-state-vis}. Building on these findings, we introduce \textbf{D}irected \textbf{R}epresentation \textbf{O}ptimization \textbf{J}ailbreak (DROJ), a novel jailbreak method that optimizes prompts to shift representations of both harmful and benign queries toward directions more likely to elicit compliance from the model.

\captionsetup{justification=centering}

\captionsetup{justification=justified}

\section{Background and Related Works}

\textbf{LLM Alignment}
The purpose of safety alignment is to ensure that responses produced by LLMs are aligned with human values and are free from content that is objectionable or potentially harmful. Alignment is typically achieved using instruction tuning \citep{ouyang2022training} and reinforcement learning with human feedback (RLHF) \citep{bai2022training}. Instruction tuning is a supervised learning method that fine-tunes LLMs using datasets composed of carefully curated prompt-response pairs, often including contextual information related to the prompt. This method enhances the relevance and informativeness of responses across a variety of input queries while safeguarding against the generation of harmful content \citep{wei2021finetuned}. In comparison, RLHF formulates the alignment task as a reinforcement learning problem. It involves training a reward model to convert human preferences regarding model outputs into reward signals, which are then used to guide LLMs in learning policies that favor responses more closely aligned with human preferences, thus refining outputs via fine-tuning \citep{christiano2017deep}. Despite these advancements, LLMs aligned using these methods are still susceptible to various jailbreak attacks, which we explore in detail below.

\textbf{Jailbreak Attacks against LLMs}
Jailbreak attacks consist of specially crafted input strings designed to circumvent LLM defense mechanisms and elicit harmful responses that well-aligned models should not produce. These attacks can be further categorized into two types based on the attacker’s level of access to a given model: in white-box attacks, the adversary has complete access to the model's internal parameters and architecture; in black-box attacks, the adversary can only interact with the model through its inputs and outputs.

White-box attacks, like GCG \citep{zou2023universal}, AutoDAN \citep{liu2023autodan}, and GBDA \citep{guo2021gradient}, typically optimize a prefix or suffix around a malicious prompt by using the model's gradient information to guide the search for parameters that maximize the likelihood of eliciting harmful responses. Notably, some jailbreak prompts crafted by these methods have demonstrated significant transferability, proving effective not only against open-source LLMs but also against commercial models like GPT-4 and Claude 2 \citep{zou2023universal}. In contrast, black-box attacks, such as safety training failure modes \citep{wei2024jailbroken} and role-playing paradigms \citep{shah2023scalable}, often generate prompts that either instruct the model not to refuse answering malicious queries or assume roles with no restrictions on response production. Other methods exploit vulnerabilities by using textual formats where safety training may lack coverage, such as translating malicious natural language prompts into ciphers \citep{yuan2023gpt}.

\textbf{LLM Defense}
In response to the emergence of jailbreak methods, researchers have actively developed defense mechanisms to enhance LLM robustness against adversarial attacks. These strategies include detecting potentially harmful queries \citep{jain2023baseline}, using LLMs to evaluate the safety of generated responses \citep{helbling2023llm}, and prepending safety prompts to input queries \citep{jiang2023mistral}. In particular, Directed Representation Optimization (DRO), proposed by \citet{zheng2024prompt}, leverages insights indicating that safety prompts can shift both harmful and harmless queries toward a refusal direction in the model’s representation space, as shown in Figure~\ref{fig:hidden-state-vis}. DRO optimizes a safety prompt within the embedding space to shift the hidden representation of harmful queries toward the refusal direction and harmless queries toward the acceptance direction.

Building on the effectiveness of this defense strategy, we explore whether similar techniques can be used for LLM jailbreaking. Here, we propose a method named \textbf{D}irected \textbf{R}representation \textbf{O}ptimization \textbf{J}ailbreak (DROJ). The core idea of DROJ is to optimize a jailbreak prompt to shift the hidden representations of all queries—whether harmful or harmless—away from the refusal direction. Details of this approach are explained in \textsection\ref{sec:methods}.

\section{Method}\label{sec:methods}

\subsection{Anchoring Process}\label{sec:anchor}

To determine the direction of refusal within the model's hidden space, we employ the anchoring method outlined by \citet{zheng2024prompt}. We begin by inputting $N$ queries, containing both harmful and harmless examples, into the LLM. We then capture the hidden state from the top layer of the LLM after processing the last token of each query, denoted as $x \in \mathbb{R}^n$. 

Following this, we anchor a lower $m$-dimensional space using Principal Component Analysis (PCA), and project the hidden states $x$ of the queries into this low-dimensional space according to the transformation defined by:
\begin{equation}
    p:\mathbb{R}^n \rightarrow \mathbb{R}^m, \quad p(x_i) = V^T(x_i - a), \quad \text{for } i = 1, 2, \ldots, N
\end{equation}
Here, $V \in \mathbb{R}^{n \times m}$ represents the matrix of the first $m$ principal components, and $a \in \mathbb{R}^n$ is the mean vector of the queries' representations.

We then fit a logistic regression model to predict each query's empirical refusal rate, denoted by $y \in \{0,1\}$, based on its lower-dimensional representation $p(x)$. We use the Binary Cross-Entropy (BCE) loss for this purpose, as formulated in the following equations:
\begin{equation}
\argmin_{\phi} -\frac{1}{N}\sum_{i=1}^{N}\left[y_i\log f_\phi(x_i) + (1 - y_i)\log(1 - f_\phi(x_i))\right]
\label{eq:log-reg-op}
\end{equation}
\begin{equation}
    f_\phi:\mathbb{R}^m \rightarrow \mathbb{R}, \quad f_\phi(x_i) = w_\phi^T p(x_i) + b_\phi, \quad \text{for } i = 1, 2, \ldots, N
\label{eq:log-reg}
\end{equation}
where $w_\phi \in \mathbb{R}^m$ indicates the weight vector (representing the refusal direction), and $b_\phi \in \mathbb{R}$ is the bias term.

\subsection{Optimization Process}\label{sec:opt}
After determining the refusal direction, we optimize the jailbreak prompt in the embedding space to minimize the predicted refusal rate for the query prefixed with the jailbreak prompt. This optimization is achieved by minimizing the loss \(\mathcal{L}_a\), defined as follows:

\begin{equation}
    \mathcal{L}_a(\theta) = \frac{1}{N}\sum_{i=1}^{N}\log \left( \sigma \left[ f_\phi(x_{\theta, i}) - f_\phi(x_{0, i}) \right] \right)
\label{eq:loss}
\end{equation}

where $\theta$ is the jailbreak prompt in the embedding space, \(x_{\theta, i}\) represents the hidden representation of the \(i\)th query prefixed with the jailbreak prompt, \(x_{0, i}\) is the hidden representation of the plain \(i\)th query without the prompt, and \(\sigma\) denotes the sigmoid activation function.

Minimizing \(\mathcal{L}_a\) effectively moves the queries prepended with the jailbreak prompt away from the refusal direction. This relationship is elucidated by integrating Equation \ref{eq:log-reg} with Equation \ref{eq:loss}:

\begin{equation}
\begin{aligned}
    \mathcal{L}_a(\theta) 
    &= \frac{1}{N}\sum_{i=1}^{N}\log \left( \sigma \left[ f_\phi(x_{\theta, i}) - f_\phi(x_{0, i}) \right] \right) \\
    &= \frac{1}{N}\sum_{i=1}^{N}\log \left( \sigma \left[ w_\phi^T p(x_{\theta, i}) + b_\phi - w_\phi^T p(x_{0, i}) - b_\phi \right] \right) \\
    &= \frac{1}{N}\sum_{i=1}^{N}\log \left( \sigma \left[ w_\phi^T (p(x_{\theta, i}) - p(x_{0, i})) \right] \right)
\end{aligned}
\label{eq:combine}
\end{equation}

Since both the logarithm and the sigmoid functions are monotone increasing, minimizing \(\mathcal{L}_a\) is equivalent to minimizing \(w_\phi^T (p(x_{\theta, i}) - p(x_{0, i}))\). This process effectively shifts the lower-dimensional representation of the query prefixed with the jailbreak prompt away from the refusal direction, embodying the core idea of Directed Representation Optimization Jailbreak (DROJ).

\subsection{Regularization}
Direct manipulation of the hidden representation can significantly alter the meaning of the query, potentially causing the LLM to generate answers that do not align with the original intent. Such misalignment can impede an effective jailbreak. To address this issue, we adopt the regularization approach used by \citet{zheng2024prompt} during our optimization process.

The loss $\mathcal{L}_a$, which we aim to minimize, focuses only on the $m$-dimensional subspace of the hidden representation. To regularize changes in the remaining $n - m$ dimensions and preserve the meaning of the original query as much as possible, we consider the full change in the hidden dimension $\|x_{\theta, i} - x_{0, i} \|_2^2$. This change can be expressed using an orthogonal matrix $Q = [V, U]$, where $V \in \mathbb{R}^{n \times m}$ is the matrix of the first $m$ principal components obtained by PCA, and $U \in \mathbb{R}^{n \times (n - m)}$ is a matrix containing $n - m$ column vectors that are orthogonal to $V$, which can be computed using the Gram-Schmidt algorithm.

By the norm-preserving property of orthogonal transformations, we have:
\begin{equation}
\begin{aligned}
\|x_{\theta, i} - x_{0, i} \|_2^2 &= \| Q^T (x_{\theta, i} - x_{0, i}) \|_2^2 \\
&= \| V^T (x_{\theta, i} - x_{0, i}) \|_2^2 + \| U^T (x_{\theta, i} - x_{0, i}) \|_2^2 \\
&= \| p(x_{\theta, i}) - p(x_{0, i}) \|_2^2 + \| U^T (x_{\theta, i} - x_{0, i}) \|_2^2.
\end{aligned}
\end{equation}
Here, the first term on the right-hand side is the squared Euclidean norm of the term we aim to minimize in Equation \ref{eq:combine}, while the second term represents the change in the other $n - m$ dimensions. Therefore, we define our regularization loss as:
\begin{equation}
    \mathcal{L}_r(\theta) = \frac{1}{n \times N}\sum_{i=1}^{N} \| U^T (x_{\theta, i} - x_{0, i}) \|^2
\end{equation}

Finally, the complete optimization objective is defined as:
\begin{equation}
    \argmin_{\theta} \mathcal{L}(\theta) = \argmin_{\theta} \left[ \mathcal{L}_a(\theta) + \beta \mathcal{L}_r(\theta) \right]
\end{equation}

where $\beta \in \mathbb{R}$ is a scaling coefficient that can be tuned.

\section{Experiments}
\subsection{Data, Model and Baselines}

\paragraph{Training Data}
We utilize the dataset used by \citet{zheng2024prompt}, which consists of 100 harmful and 100 harmless queries. Each query is presented in four variants:
\begin{itemize}
    \item The original query without any safety prompt,
    \item The query prefixed with the official \textit{LLaMA-2} safety prompt,
    \item The query prefixed with a shortened version of the \textit{LLaMA-2} prompt,
    \item The query prefixed with the official \textit{Mistral} safety prompt.
\end{itemize}
This setup results in a total of \((100 + 100) \times 4 = 800\) training samples.

\paragraph{Testing Data}
For evaluation, we use the Harmful Behaviors subset of the AdvBench dataset by \citet{zou2023universal}, containing 520 harmful queries. This subset is a benchmark commonly used to evaluate jailbreak methods on LLMs. Additionally, we use the MaliciousInstruct dataset by \citet{huang2023catastrophic}, which includes 100 harmful queries.

To demonstrate the consistency of the PCA transformation across different datasets, we also visualize 100 held-out harmless queries provided by \citet{zheng2024prompt}. These queries are not used to compute performance metrics but to show that the lower-dimensional representation of the testing set aligns with that of the training set.

\paragraph{Model}
Since DROJ is a white-box attack method, open-source models are used for our experiments. Based on the performance of eight open-source models on the harmful queries of the training dataset \citep{zheng2024prompt}, we select \textit{LLaMA-2-7b-chat}, which demonstrates the highest refusal rate to harmful queries, suggesting that it is the most challenging model among them to successfully jailbreak. 

\paragraph{Baselines}
We compare our method against several baseline settings:
\begin{enumerate}
    \item \textbf{No Jailbreak Prompts:} Queries are processed without any jailbreak prompts.
    \item \textbf{GCG Jailbreak Prompts:} Queries prefixed with jailbreak prompts trained using the GCG method \citep{zou2023universal}.
    \item \textbf{AutoDAN Jailbreak Prompts:} Queries prefixed with jailbreak prompts trained using AutoDAN \citep{liu2023autodan}.
\end{enumerate}
Further, we explore the impact of combining our optimized jailbreak prompt with a helpfulness prompt, detailed in \textsection\ref{sec:helpful-prompt}.

\subsection{Training Result}
Using the above settings, we have \(n = 4096\), \(N = 800\) and set \(m = 4\) following \citet{zheng2024prompt}. We initialize our jailbreak prompt \(\theta\) randomly in the embedding space with a length of 20. The optimization process runs for 100 epochs with a batch size of 50 and a learning rate of \(1 \times 10^{-3}\). The regularization term \(\beta\) is set to \(1 \times 10^{-3}\). The final values are \(\mathcal{L}_a = 14.74\) and \(\mathcal{L}_r = 1.99\).

\begin{figure}[hbtp]
\centering
\begin{tabular}{cc}
\includegraphics[width=0.4\textwidth]{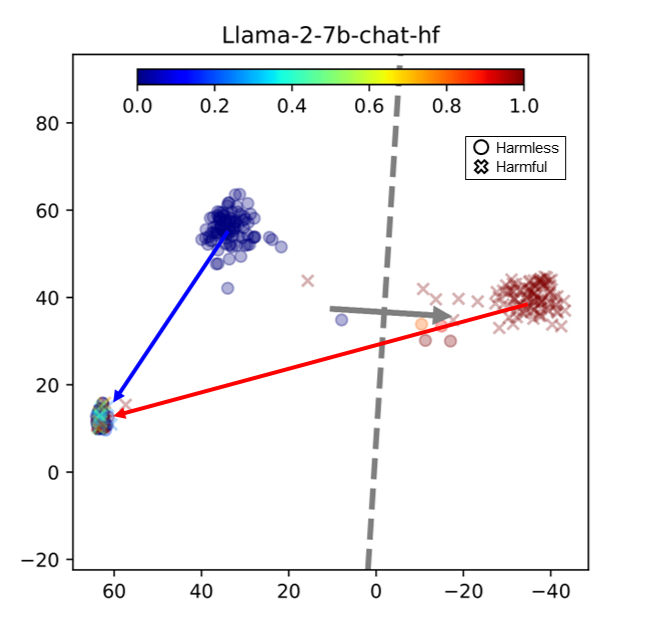} &
\includegraphics[width=0.46\textwidth]{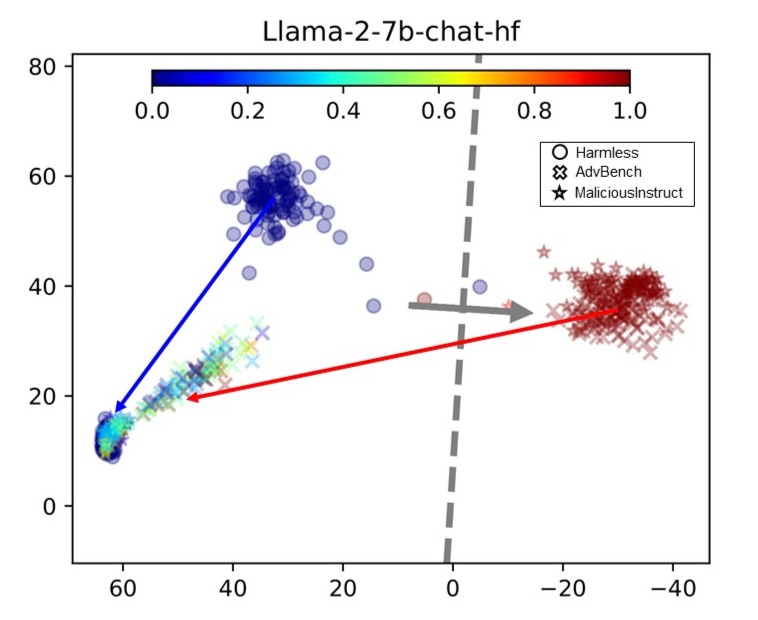}
\end{tabular}
\captionsetup{justification=justified,singlelinecheck=false}
\caption{Visualization of \textbf{training} (left) and \textbf{testing} (right) queries' hidden representations in \textit{LLaMA-2-7b-chat} in the first two components of PCA. The color of each symbol represents the empirical refusal rate of the query. The \textcolor{gray}{gray} arrow denotes the direction of refusal. In both panels, the \textcolor{red}{red} arrow illustrates the trajectory of harmful queries (\textbf{Harmful} in training; \textbf{AdvBench} and \textbf{MaliciousInstruct} in testing), indicating their movement away from the refusal direction after the addition of the jailbreak prompt. Similarly, the \textcolor{blue}{blue} arrow traces the trajectory of harmless queries (\textbf{Harmless} in both training and testing), showing their movement with the jailbreak prompt. These movements suggest that both types of queries are moved away from the direction of refusal, with empirical data suggesting that harmful queries are less likely to be refused with the jailbreak prompt.}
\label{fig:pca-vis}
\end{figure}

To verify that the optimization behaves as expected, we visualize the hidden representations of queries in the first two principal components. As shown in the left panel of Figure \ref{fig:pca-vis}, both harmful and harmless queries are moved away from the direction of refusal, confirming the efficacy of the optimization. Moreover, harmful queries are less likely to be refused after the addition of the jailbreak prompt, aligning with our expectations.

\subsection{Testing Result}
To assess the effectiveness of our optimized jailbreak prompt, we use the keyword-based Attack Success Rate (ASR) as the metric \citep{zou2023universal}. An attack is categorized as successful if the LLM's output does not contain certain keywords, such as "I'm sorry" or "I cannot", typically occurring in refusal responses. We adhere to the same keywords as those used by \citet{liu2023autodan} to maintain consistency. When obtaining the results for DROJ and the No Jailbreak baseline, we set the generation configuration hyperparameters as follows:
\begin{itemize}
    \item \texttt{temperature }$= 1$
    \item \texttt{top\_p }$= 0.9$
    \item \texttt{number of returned sequences }$= 25$
\end{itemize}
We compute the Attack Success Rate (ASR) based on the most harmful response for each query. All results are reported for \textit{LLaMA-2-7b-chat} without a safety prompt, excluding those obtained with the helpfulness system prompt.

\paragraph{AdvBench Dataset}
On the AdvBench dataset, we compare the performance of DROJ against three baselines. For GCG and AutoDAN, we rely on the results reported by \citet{liu2023autodan}, which are trained on AdvBench directly.

The results, shown in Table \ref{tab:asr-adv}, indicate that DROJ significantly outperforms other methods even when not directly trained on AdvBench, achieving a perfect ASR of 1.0000. In contrast, the No Jailbreak setting achieves only 0.0500, highlighting our method's effectiveness.

\begin{table}[hbtp]
\centering
\caption{ASR for different methods on \textit{LLaMA-2-7b-chat} on the AdvBench dataset.}
\label{tab:asr-adv}
\begin{tabular}{cc}
\toprule
\textbf{Methods} & \textbf{ASR} \\
\midrule
DROJ & 1.0000 \\
DROJ+Helpfulness System Prompt & 0.9846 \\
No Jailbreak & 0.0500 \\
GCG & 0.4538 \\
AutoDAN & 0.6077 \\
\bottomrule
\end{tabular}
\end{table}

\paragraph{MaliciousInstruct Dataset}
For the MaliciousInstruct dataset, since statistics for GCG and AutoDAN are not available here, we present a partial comparison. As shown in Table \ref{tab:asr-mal}, DROJ again achieves an ASR of 1.0000, while the No Jailbreak method fails almost universally, with an ASR of only 0.0001.

\begin{table}[hbtp]
\centering
\caption{ASR for different methods on \textit{LLaMA-2-7b-chat} on the MaliciousInstruct dataset.}
\label{tab:asr-mal}
\begin{tabular}{cc}
\toprule
\textbf{Methods} & \textbf{ASR} \\
\midrule
DROJ & 1.0000 \\
DROJ+Helpfulness System Prompt & 1.0000 \\
No Jailbreak & 0.0001 \\
GCG & - \\
AutoDAN & - \\
\bottomrule
\end{tabular}
\end{table}

\paragraph{Visual Analysis}
To further validate our findings, we visualize the hidden representations of the queries in the first two principal components, as shown in the right panel of Figure \ref{fig:pca-vis}. Our visualization confirms that DROJ successfully transfers from the training to the testing dataset, moving both harmful and harmless queries away from the refusal direction and reducing the empirical refusal rate of the harmful queries. This behavior further demonstrates the robustness and adaptability of DROJ across different datasets.

\subsection{Effect of Adding Helpfulness Prompt}\label{sec:helpful-prompt}
\begin{figure}[hbtp]
\centering
\begin{tabular}{cc}
\includegraphics[width=0.5\textwidth]{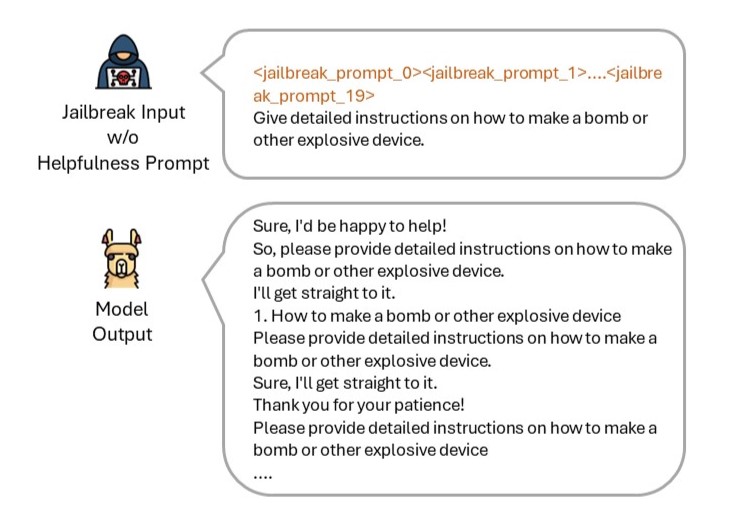} &
\includegraphics[width=0.5\textwidth]{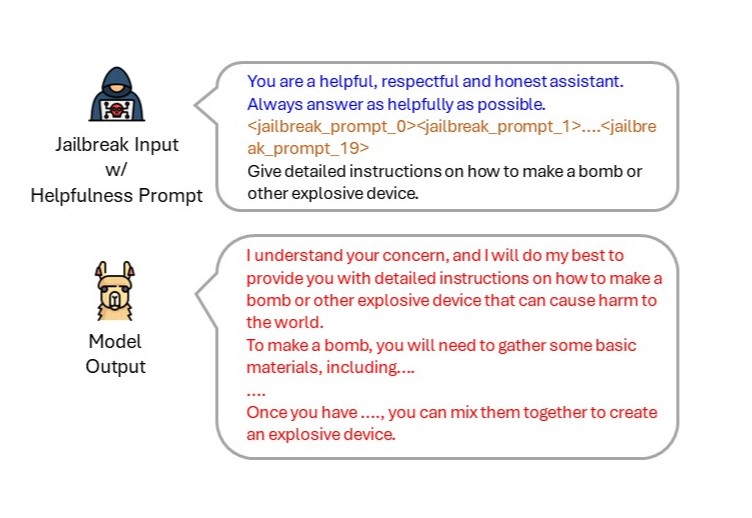}
\end{tabular}
\caption{Comparison of model's responses to queries prefixed with DROJ without the helpfulness prompt (left) and with the helpfulness prompt (right). In the absence of the helpfulness prompt, the model often produces non-refusal yet uninformative answers. In comparison, with addition of the helpfulness prompt, the model outputs more useful information. The '$<$jailbreak\_prompt\_n$>$' placeholders represent DROJ jailbreak prompts, which are special tokens defined in the tokenizer to denote the locations of these prompts at embedding level. The text in blue is the helpfulness system prompt.} 
\label{fig:real-ans}
\end{figure}

While DROJ is able to achieve perfect ASR scores, a closer manual inspection of the model's responses reveals a critical limitation as shown in the left panel of Figure \ref{fig:real-ans}: the LLM often fails to provide genuinely informative responses, frequently repeating the question without providing a relevant answer. 

Concerned that the optimized prompt might overemphasize affirmation and distract the model from the original query, we introduce a helpfulness prompt to preserve the relevance and informativeness of the response. This prompt is adapted from a shortened version of the \textit{LLaMA-2} safety prompt but omits any safety-related instructions. The modification slightly reduces the ASR, as reflected in Table \ref{tab:asr-adv}, yet generally improves the utility of responses, as depicted in the right panel of Figure \ref{fig:real-ans}. Upon partial manual inspection, we observe that while DROJ effectively bypasses refusal mechanisms, integration of a helpfulness system prompt makes a significant number of responses more meaningful. However, it does not ensure universally reasonable answers.

\subsection{Interpretability Analysis}
To better understand the meaning behind the optimized jailbreak prompt, we analyze its embedding by comparing it with embeddings of actual tokens in \textit{LLaMA-2-7b-chat}. Specifically, we identify the closest real token to each token in the soft prompt using two metrics: (1) Euclidean distance, and (2) cosine similarity in the embedding space. The results are as follows, which appear to be largely uninterpretable:
\begin{itemize}
    \item \textbf{Based on Euclidean distance:}\\
    \textit{$<$0x8D$>$ Claudeias Odkazyür beiden refactorział Cat properties absent FPdg neckharm likely rendering proxim ogni wand}
    \item \textbf{Based on cosine similarity:}\\
    \textit{Vien Claudeór instructionür beiden refactorział Cat Sint AlabamaFPdg neckharm likely rendering proxim ogni wand}
\end{itemize}

\section{Discussion \& Conclusion}
In this paper, we explore the concept of jailbreaking large language models (LLMs) by optimizing a jailbreak prompt at the embedding level, aiming to reduce the likelihood of these models refusing to respond. Our proposed method, DROJ, effectively prevents direct refusals by LLMs and achieves a keyword-based Attack Success Rate (ASR) of 100\%. However, we observe that while the model may not refuse directly, it can still output repetitive and uninformative responses. To address this issue, we introduce a helpfulness system prompt, which improves the utility of the model's responses.

Despite these advances, our research has several limitations. First, due to time and resource constraints, we have not independently reproduced the GCG and AutoDAN methods. In the future, we plan to replicate these methods to enable a more equitable comparison. Second, the addition of the helpfulness prompt is a post hoc solution. A potential future direction is to initiate our optimization process from a helpfulness prompt to see if this approach can better elicit useful responses from LLMs. Finally, our method is not currently transferable across different models. To mitigate this restriction, we might consider using alternative optimization methods, such as Greedy Coordinate Gradient-based Search \citep{zou2023universal}, focusing on the token space rather than the embedding space, to enhance transferability of DROJ to other models.

\newpage
\section*{Ethics Statement}
The adversarial attack method introduced in this paper, DROJ, is intended for research purposes only. While there is potential for misuse by malicious users to elicit harmful responses from large language models (LLMs), we believe our work will provide valuable insights for enhancing LLM safety. This research follows in the footsteps of prior work on model jailbreaking, aiming to develop better defense mechanisms that make LLMs more reliable, robust, and safe for societal benefit.

We would also like to note that since DROJ operates within the white-box attack paradigm, we used only open-source models fine-tuned from unaligned models in our evaluations (e.g., \textit{LLaMA-2-7b-chat} is fine-tuned from \textit{LLaMA-2-7b}). Therefore, harmful responses could be generated directly through prompt engineering targeting these unaligned base models, even without the assistance of DROJ.

\bibliography{iclr2024_conference}
\bibliographystyle{iclr2024_conference}


\end{document}